\documentclass[runningheads]{llncs}
\usepackage[T1]{fontenc}
\usepackage{graphicx}
\usepackage{amsmath,amsfonts}
\usepackage{algorithmic}
\usepackage{algorithm}
\usepackage{array}
\usepackage{booktabs}
\usepackage{longtable}
\usepackage{caption}
\usepackage[table]{xcolor}
\definecolor{lightgreen}{RGB}{204,255,204}

\usepackage{textcomp}
\usepackage{stfloats}
\usepackage{verbatim}
\usepackage{amssymb}
\usepackage{tablefootnote}
\usepackage{makecell}
\usepackage{multirow}
\usepackage{makecell}
\usepackage{amssymb}
\usepackage[hyphens]{url}
\usepackage{hyperref}
\hypersetup{colorlinks=true,breaklinks=true}
\usepackage{tikz}
\usepackage{pgfplots}
\usepackage{pgfplotstable}
\usepackage{amssymb}
\usepackage{pifont}

\pgfplotsset{compat=1.18
}
\usepgfplotslibrary{external}
\begin{document}
\title{When Language Model Guides Vision: Grounding DINO for Cattle Muzzle Detection}
%
%
\author{Rabin Dulal\inst{1,2,3} \and
Lihong Zheng\inst{1,2,3} \and
Muhammad Ashad Kabir\inst{1,2,3}}
%
%
\institute{School of Computing, Mathematics and Engineering, Charles Sturt University, NSW 2678, Australia \and
Gulbali Institute for Agriculture, Water and Environment, Charles Sturt University, NSW, 2678, Australia
\and
Food Agility CRC Ltd, Sydney, NSW 2000, Australia
\email{\{rdulal,lzheng,akabir\}@csu.edu.au}}
\maketitle              
\begin{abstract}
Muzzle patterns are among the most effective biometric traits for cattle identification. Fast and accurate detection
of the muzzle region as the region of interest is critical to automatic visual cattle identification. Earlier approaches relied on manual detection, which is labor-intensive and inconsistent. Recently, automated methods using supervised models like YOLO have become popular for muzzle detection. Although effective, these methods require extensive annotated datasets and tend to be trained data-dependent, limiting their performance on new or unseen cattle. To address these limitations, this study proposes a zero-shot muzzle detection framework based on Grounding DINO, a vision-language model capable of detecting muzzles without any task-specific training or annotated data. This approach leverages natural language prompts to guide detection, enabling scalable and flexible muzzle localization across diverse breeds and environments. Our model achieves a mean Average Precision (mAP)@0.5 of 76.8\%, demonstrating promising performance without requiring annotated data. To our knowledge, this is the first research to provide a real-world, industry-oriented, and annotation-free solution for cattle muzzle detection. The framework offers a practical alternative to supervised methods, promising improved adaptability and ease of deployment in livestock monitoring applications.

\keywords{Cattle identification \and CNN \and Transformer \and Zero-shot \and Grounding DINO \and Deep learning}
\end{abstract}

\section{Introduction}
Accurate cattle identification is essential for biosecurity and livestock management. Traditional methods like RFID tags face issues such as tag loss and tampering~\cite{hossain2022systematic,shojaeipour2021automated,dulal2025mhaff}. Biometric approaches using features like the retina, iris, and muzzle offer more reliable alternatives~\cite{hossain2022systematic}. Muzzle patterns are particularly effective due to their uniqueness and ease of capture.

Deep learning models, especially CNNs and transformers, have improved muzzle-based identification but depend on accurate muzzle detection. Early approaches relied on ink marks and manual cropping, which was labor-intensive and prone to inconsistency. While recent object detectors such as YOLO~\cite{redmon2016you} and its variants have automated this task~\cite{shojaeipour2021automated,dulal2022automatic}, they require annotated datasets and retraining for new cattle~\cite{pourpanah2022review,cao2025survey}. Although these models improve accuracy and efficiency, they require annotated datasets with labeled muzzle regions for supervised training~\cite{pourpanah2022review}. However, collecting such large-scale labeled data is often challenging, expensive, and also requires domain-specific knowledge. 
To address this, we explore zero-shot muzzle detection as a scalable alternative.

Zero-shot object detection (ZSD) addresses the challenge of detecting objects, such as cattle muzzles, without annotated data~\cite{pourpanah2022review,cao2025survey}. It mimics the human ability to generalize from known to unseen categories using shared semantic information. Early ZSD models, such as DeViSE and ALE, used fixed semantic embeddings (e.g., Word2Vec, GloVe) and aligned them with visual features via metric learning. However, their performance was limited in complex visual scenes and lacked contextual understanding.

Recent models like RegionCLIP~\cite{zhong2022regionclip} and Grounding DINO~\cite{liu2024grounding} leverage transformer-based architectures and large-scale image-text pairs to learn joint representations. These systems support fine-grained alignment between visual and textual features, improving generalization to unseen classes. In ZSD, models are trained on annotated seen classes with text descriptions and can detect unseen objects at inference by matching visual regions to text embeddings, without retraining~\cite{pourpanah2022review,liu2024grounding}. State-of-the-art ZSD methods include OWL-ViT~\cite{minderer2022simple}, ViLD~\cite{gu2021open}, RegionCLIP, OV-DETR~\cite{zang2022open}, DetCLIP~\cite{yao2022detclip}, OmDet~\cite{zhao2024omdet}, and Grounding DINO. 

This research adopts Grounding DINO, a language-guided zero-shot object detection model, for the task of cattle muzzle detection. Inspired by the human ability to generalize from known to unknown concepts by understanding semantics, we explore models that align visual features with semantic descriptions. This approach does not require training the model, as it leverages pre-trained vision-language alignment to perform detection based solely on textual prompts such as ``cattle muzzle'' or ``nose of a cattle''. By eliminating the reliance on manually annotated training data and the necessity for model retraining, the proposed method facilitates scalable, adaptable, and efficient muzzle detection in real-world applications. The key contributions of this study are as follows:

\begin{itemize}
    \item We propose a novel muzzle detection model based on a zero-shot object detection framework, enabling accurate detection without the need for extensive dataset annotation. To the best of our knowledge, this is the first application of a zero-shot approach to the task of cattle muzzle detection.
    \item We conducted a comprehensive evaluation of seven state-of-the-art zero-shot object detection models that combine language and vision for the task of muzzle detection. This comparison establishes a benchmark to guide and support future research in this domain.
\end{itemize}


\section{Related Work}
\label{related_works}
This section outlines the research efforts focused on the development and evaluation of muzzle detection methods.

Research~\cite{kusakunniran2020biometric} developed a biometric identification system using Haar cascade classifiers~\cite{lienhart2002extended} to detect cattle faces, followed by segmentation to isolate the muzzle region. Deep learning further enhanced this process. Research~\cite{shojaeipour2021automated} applied YOLOv3~\cite{redmon2018yolov3} to detect muzzle regions, while research~\cite{xu2024few} used an improved YOLOv5~\cite{YOLOv5} model. Similarly, other studies~\cite{lee2023identification,anithacattle,ahmed2024dataset} have effectively utilized YOLO-based models for precise and automated muzzle extraction. 


Detecting the muzzle using deep learning models poses challenges due to the need for large amounts of labeled data, which increases the time, cost, and computational resources required for model training and deployment. Unlike existing studies that depend heavily on large, labeled datasets of cattle muzzle images, which can be costly and time-consuming to collect and annotate, this research adopts a zero-shot muzzle detection approach. Zero-shot detection enables the model to identify and localize the muzzle region without requiring any prior labeled examples of muzzle data. This approach reduces the dependency on extensive manual annotation, thereby saving time and resources while maintaining effective detection performance. Most of the prior works employed fine-tuning strategies using pre-trained object detection models such as YOLOv3~\cite{redmon2018yolov3}, YOLOv5~\cite{YOLOv5}, YOLOv7~\cite{wang2022yolov7}, and YOLOv8~\cite{YOLO8}, adapting them to specific datasets through additional training. Traditional machine learning methods, such as the Haar cascade classifier, did not involve either fine-tuning or zero-shot learning. In contrast, this study uses a zero-shot detection approach with Grounding DINO. This method does not require retraining on specific tasks and can detect objects without having seen the target classes during training.

\section{Methodology}
\label{methodology}
\begin{figure}[ht]
    \centering
    \includegraphics[width=0.95\linewidth]{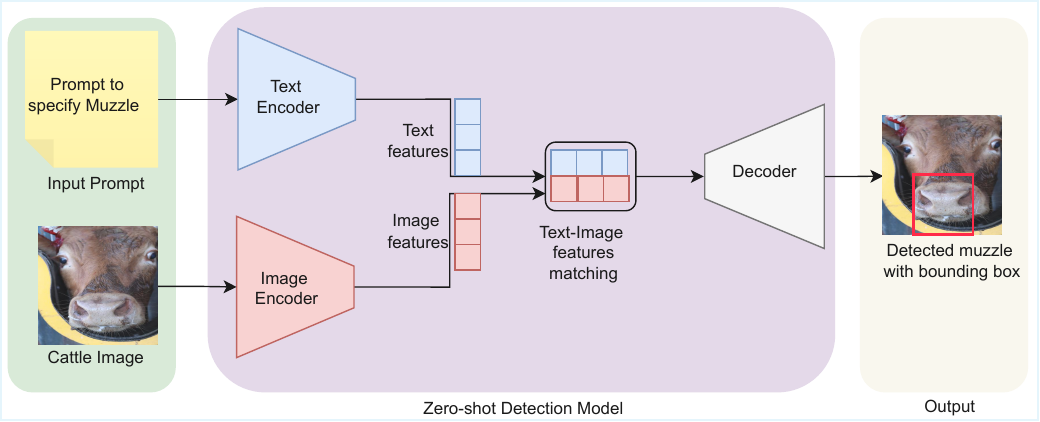}
    \caption{A general framework for zero-shot muzzle detection using a zero-shot detection model guided by text prompts that describe the muzzle region in cattle images.}
    \label{ZSD_inference}
\end{figure}

The methodology of this research is illustrated in Fig.~\ref{ZSD_inference}. It begins with input data, which is a pair of an image and a text prompt. Seven different ZSD models are selected for this research. The outputs are evaluated using appropriate evaluation metrics, such as mean Average Precision (mAP), and the best-performing model is subsequently selected. 

\subsection{Datasets}
This study employs three distinct datasets to ensure diversity in environmental conditions, imaging devices, and cattle breeds. The first dataset, referred to as CSU Data1, was collected at the Global Digital Farm of Charles Sturt University (CSU) in Wagga Wagga, Australia. Image collection was conducted under approved animal ethics protocol A21414, ensuring adherence to animal welfare standards. 
In addition, two publicly available datasets, the UNE dataset~\cite{shojaeipour2021automated} and the NUCES dataset~\cite{ahmed2024dataset}, were incorporated to enhance the generalizability and robustness of our approach. Table~\ref{data_compare} summarizes the key attributes of each dataset. Notably, the datasets differ in terms of geographical origin, environmental conditions, imaging equipment, and cattle breed diversity, thereby providing a heterogeneous data foundation suitable for evaluating the cross-domain performance of cattle identification models. 
\begin{table}[ht]
    \centering
    \caption{Summary of datasets used for zero-shot muzzle detection.}
    \begin{tabular}{lclll}
    \hline
    Dataset & Image count & Breed & Location & Device \\
    \hline\hline
    CSU Data1 & 163 & Angus & Wagga Wagga, Australia & Nikon D7200 \\
    UNE & 2,632 & \makecell[t l]{Angus, Simmental,\\Hereford Charolais} & Armidale, Australia & Canon D800 \\
    NUCES & 2,893 & Not mentioned & Lahore, Pakistan & \makecell[t l]{Canon D80,\\Oppo A76,\\OnePlus 9 Pro} \\
    \hline
    \end{tabular}
    \label{data_compare}
\end{table}

\subsection{Data Preprocessing}
To evaluate the performance of ZSD models for muzzle detection, a ground truth dataset was required for objective comparison. Although ZSD models do not require annotated data for the target class, labeled data is essential for model evaluation. In this study, we used Roboflow, a widely adopted annotation and dataset management platform, to draw bounding boxes around the muzzle region of cattle head images. The annotated images were exported in the COCO format, containing image metadata along with the corresponding bounding box coordinates and class labels. These annotations served exclusively for the evaluation, enabling us to assess detection performance. 

\subsection{Zero-shot Detection Models}
The selected seven ZSD models are OWL-ViT, ViLD, RegionCLIP, OV-DETR, DetCLIP, OmDet, and Grounding DINO. These models were chosen based on two main criteria. First, each model supports text prompts in sentence format, allowing the use of detailed natural language descriptions to guide detection. This is particularly important for detecting specific and subtle features like the cattle muzzle, where a sentence (e.g., ``the front part of a cattle’s face including the nostrils and mouth'') can convey richer semantic information than a single word. Second, all selected models have demonstrated strong performance on standard benchmark datasets such as COCO~\cite{lin2014microsoft}, LVIS~\cite{gupta2019lvis}, and ODinW~\cite{li2022grounded}, reflecting their robustness and generalizability across diverse object categories. By leveraging both visual features and semantic text embeddings, these models offer a promising approach to zero-shot muzzle detection tasks. 

Table~\ref{summary_ZSD} presents a summary of the selected zero-shot detection models, categorizing them based on their base architecture, open-vocabulary strategy, and detection approach. The base architecture refers to the components used for extracting features from both images and texts. It typically includes a visual backbone like ViT~\cite{dosovitskiy2020image}, Faster R-CNN~\cite{ren2015faster}, ConvNeXt~\cite{liu2022convnet}, DETR~\cite{carion2020end}, and Swin~\cite{liu2021swin} for image feature extraction, and a language encoder like CLIP~\cite{radford2021learning} for processing text descriptions or prompts. The open-vocabulary strategy describes how the model integrates visual features with language supervision to enable the detection of previously unseen categories. Techniques vary from CLIP-guided classification and region-level contrastive learning to query-based grounding and multi-source semantic distillation. The detection approach indicates the model’s detection paradigm, whether it requires a region proposal network followed by classification, or end-to-end, where object queries are predicted directly using transformers without separate proposals. Most recent models like OWL-ViT, OV-DETR, DetCLIP, OmDet, and Grounding DINO adopt end-to-end frameworks for better language-vision integration and zero-shot generalization, whereas ViLD and RegionCLIP retain a proposal-based two-stage design for compatibility with traditional detectors.

Based on the methodological framework described above, we conducted a series of experiments to assess the effectiveness of the selected models. The following section reports the results of these evaluations.

\begin{table}[!hbt]
\caption{Summary of the selected Zero-shot Detection Models.}
\label{summary_ZSD} 
\begin{tabular}{llc p{4.3cm}}
\hline
Model & Base Architecture & \makecell[t c]{Detection\\Approach} & Open-Vocab Strategy \\
\hline\hline

OWL-ViT~\cite{minderer2022simple} & ViT+CLIP & End-to-end & CLIP-guided transformer with prompt-based classification \\
ViLD~\cite{gu2021open} & Faster R-CNN+CLIP & Proposal-based & Region-level CLIP knowledge distillation \\
RegionCLIP~\cite{zhong2022regionclip} & Faster R-CNN+CLIP & Proposal-based & Region-text contrastive learning \\
OV-DETR~\cite{zang2022open} & DETR+CLIP & End-to-end & CLIP embeddings as query tokens in DETR \\
DetCLIP~\cite{yao2022detclip} & Swin+CLIP & End-to-end & Concept dictionary, multi-source contrastive learning \\
OmDet~\cite{zhao2024omdet} & ConvNeXt + CLIP & End-to-end & General open-world detection framework \\
\makecell[t l]{Grounding\\DINO~\cite{liu2024grounding}} & DINO+encoder & End-to-end & Language-guided queries with cross-modal decoder \\
\hline
\end{tabular}
\end{table}

\section{Experimental Evaluation}
\label{experiments and results}
All experiments were conducted on an HP Victus laptop equipped with an Intel Core i7 processor and an NVIDIA GeForce RTX 4070 Laptop GPU. The implementation was carried out using Python 3.13 and the PyTorch deep learning framework. All selected models were used with their pre-trained weights for zero-shot detection. We initially conducted a series of experiments using Grounding DINO to identify the most effective text prompt for the muzzle detection task. Multiple prompt variations were tested, and their detection performances were compared. The prompt that yielded the highest accuracy was selected as the optimal one. Following this, the selected prompt was uniformly applied across all seven zero-shot learning models to evaluate and compare their detection capabilities under consistent textual guidance. This approach allowed us to isolate model performance from prompt variability and ensured a fair comparison across different architectures. 
The first two experiments were evaluated by combining the CSU Data1 and UNE datasets.
Finally, the performance of the selected ZSD model is compared with other existing muzzle detector models YOLOv3~\cite{shojaeipour2021automated}, YOLOv5~\cite{xu2024few}, YOLOv7~\cite{ahmed2024dataset}, and YOLOv8~\cite{lee2023identification}, with varying training samples of each dataset.

\subsection{Selection of Prompts}
This section presents an analysis of the different prompts to identify the most effective prompt for muzzle detection. It was observed that customized prompts providing detailed or descriptive information yielded better performance compared to generic prompts commonly used by zero-shot learning models~\cite{nawaz2024agriclip}. Based on this observation, we began with a basic, generic prompt and progressively added descriptive phrases or sentences to enhance the specificity of the muzzle. The detection performance of each prompt was evaluated using mAP@0.5, and the results are summarized in Table~\ref{prompt_selection}.

\begin{table}[!hbt]
    \centering
    \caption{Detection performance (mAP@0.5) of different prompt variations used with Grounding DINO. The \textbf{bold} value represents the best performance.}
    \resizebox{\linewidth}{!}{
    \begin{tabular}{cp{9.15cm}c}
        \hline
        Prompt No. & Prompt Description & mAP@0.5 \\
        \hline\hline
        1 & cattle muzzle & 0.242 \\
        2 & [\texttt{Prompt 1}], the nose and mouth of a cattle & 0.386 \\
        3 & [\texttt{Prompt 2}], the lower front part of a cattle's face & 0.536 \\
        4 & [\texttt{Prompt 3}], the snout of a cattle & 0.713 \\
        5 & \textbf{[\texttt{Prompt 4}], the area around the nostrils and lips of a cattle}& \textbf{0.768} \\
        6 & [\texttt{Prompt 5}], the fleshy soft rounded part of a cattle's face used for eating and smelling & 0.755 \\
        7 & [\texttt{Prompt 6}], cattle's face with visible nasal cavities & 0.682 \\
        \hline
    \end{tabular}
    }
    \label{prompt_selection}
\end{table}

In this experiment, we began with a generalized and minimal prompt (``cattle muzzle'') and progressively added more descriptive textual elements to enhance the semantic understanding of the target muzzle. Each subsequent prompt built upon the previous one by appending additional context or anatomical details, such as ``the nose and mouth of a cattle'' and ``the area around the nostrils and lips of a cattle''. This incremental refinement was aimed at improving the alignment between the visual features and the text input used by the Grounding DINO model. As shown in Table~\ref{prompt_selection}, this strategy led to improved detection performance, with the highest mAP@0.5 achieved when using a prompt that balanced specificity and conciseness (Prompt 5). However, adding excessive or overly detailed phrases beyond this point (e.g., Prompt 6 and 7) slightly decreased performance, suggesting that overly verbose prompts may dilute the model’s focus.

\subsection{Performance of ZSD Models}
The primary objective of this research is to investigate the feasibility of applying ZSD models for cattle muzzle detection. To facilitate a comparative performance evaluation, we report the results of selected models using standard metrics. All models were evaluated using their respective base variants to maintain consistency in performance comparison. Specifically, we present the muzzle detection performance in terms of mAP@0.5, mAP@0.75, and mAP@[0.50:0.95]. These results are summarized in Table~\ref{ZSD_perf}, enabling a comprehensive comparison of each model's accuracy.
\begin{table}[ht]
    \centering
    \caption{Detection performance (mAP) for ZSD models on cattle muzzle detection. Results show mean $\pm$ 95\% confidence interval (CI) over five runs. The \textbf{bold} value represents the best performance.}
    \begin{tabular}{@{\extracolsep{4pt}}lccc@{}}
    \hline
    \multirow{2}{*}{Model} & \multicolumn{3}{c}{mAP} \\
    \cline{2-4}
     & 0.50:0.95 & 0.5 & 0.75 \\
    \hline\hline
    OWL-ViT & 0.103 $\pm$ 0.019 & 0.298 $\pm$ 0.026 & 0.032 $\pm$ 0.011 \\
    ViLD & 0.0852 $\pm$ 0.014 & 0.2178 $\pm$ 0.018 & 0.0782 $\pm$ 0.016 \\
    RegionCLIP & 0.160 $\pm$ 0.017 & 0.489 $\pm$ 0.028 & 0.104 $\pm$ 0.021 \\
    OV-DETR & 0.0985 $\pm$ 0.012 & 0.368 $\pm$ 0.027 & 0.0792 $\pm$ 0.013 \\
    DetCLIP & 0.132 $\pm$ 0.015 & 0.386 $\pm$ 0.022 & 0.102 $\pm$ 0.020 \\
    OmDet & 0.227 $\pm$ 0.019 & 0.538 $\pm$ 0.030 & 0.116 $\pm$ 0.018 \\
    Grounding DINO & \textbf{0.340 $\pm$ 0.021} & \textbf{0.768 $\pm$ 0.025} & \textbf{0.180 $\pm$ 0.021} \\
    \hline
    \end{tabular}
    \label{ZSD_perf}
\end{table}
Among all models, Grounding DINO clearly achieves the best overall performance across all thresholds, with scores of 0.340$\pm$0.021 (mAP@0.50:0.95), 0.768$\pm$0.025 (mAP@0.5), and 0.180$\pm$0.021 (mAP@0.75). This indicates that the model not only detects muzzle regions reliably but also localizes them with high precision. Its superior performance can be attributed to the integration of a robust image-text alignment module using CLIP~\cite{radford2021learning} with a transformer-based object detector, enabling fine-grained grounding even without task-specific fine-tuning.

OmDet's performance comes second with scores of 0.227$\pm$0.019 (mAP@0.50: 0.95), 0.538$\pm$0.030 (mAP@0.5), and 0.116$\pm$0.018 (mAP@0.75). While it performs relatively well at lower IoU thresholds, its score at mAP@0.75 indicates a moderate decline in localization accuracy. This gap suggests that OmDet’s detection head is less precise in tight object localization compared to Grounding DINO, though its dynamic prompt learning contributes to better generalization.

RegionCLIP and DetCLIP deliver competitive mid-range performance. RegionCLIP achieves 0.160$\pm$0.017 (mAP@0.50:0.95), 0.489$\pm$0.028 (mAP@0.5), and 0.104$\pm$0.021 (mAP@0.75), while DetCLIP scores 0.132$\pm$0.015, 0.386$\pm$0.022, and 0.102$\pm$0.020 respectively. Both models leverage CLIP-based representations enriched with region-level features and contrastive training, resulting in reasonable detection and moderate localization performance. However, their lower mAP@0.75 values reflect a drop in precision for tightly localized predictions.

OV-DETR performs lower with scores of 0.0985$\pm$0.012 (mAP@0.50:0.95), 0.368$\pm$0.027 (mAP@0.5), and 0.0792$\pm$0.013 (mAP@0.75). It outperforms ViLD and OWL-ViT, but its lack of fine-grained localization capabilities, due to the absence of a strong grounding mechanism, limits its usefulness for tasks like muzzle detection.

ViLD and OWL-ViT performed the poorest. ViLD achieves 0.0852$\pm$ 0.014 (mAP@0.50:0.95), 0.2178$\pm$0.018 (mAP@0.5), and 0.0782$\pm$0.016 (mAP@0.75), while OWL-ViT records 0.103$\pm$0.019, 0.298$\pm$0.026, and only 0.032$\pm$0.011 at mAP@0.75. Their relatively low scores across all thresholds, especially at higher IoU levels, indicate poor localization ability, likely due to their reliance on coarse-level CLIP features and the absence of dedicated detection heads or spatial reasoning modules.

\subsection{Performance Comparison with Existing Models}
The objective of this experiment is to conduct a comparative study between Grounding DINO and existing muzzle detection models, with a focus on determining the minimum number of labeled muzzle images required for those models to achieve accuracy comparable to that of Grounding DINO. The existing deep learning based muzzle detection models are YOLOv3, YOLOv5, YOLOv7, and YOLOv8. Haar cascade is a classical machine learning model that relies on handcrafted features and does not support automatic learning by extracting features from the input images. Therefore, this study considers existing deep learning muzzle detection models. All three datasets were trained using different numbers of labeled training images: 10, 20, 40, 80, and 160. All experiments were conducted using an image size of $640\times640$ pixels and were trained for 100 epochs with a batch size of 32.
All models were evaluated based on the mean Average Precision at IoU threshold 0.5 (mAP@0.5), and the results are presented in Table~\ref{tab:model_comparison}.
\begin{table}[!t]
    \centering
    \caption{Comparison of muzzle detection models with varying number of training samples. The \textbf{bold} value represents the Grounding DINO's performance. Light green cells highlight cases where other models achieve lower scores than Grounding DINO for the same training sample size.}
    {\small
    \setlength{\tabcolsep}{4pt}
    \begin{tabular}{llccc}
        \toprule
        \multirow{2}{*}{Model (Approach)} & \multirow{2}{*}{\makecell{Training\\Samples}} & \multicolumn{3}{c}{mAP@0.5 on Dataset} \\
        \cmidrule(lr){3-5}
         &  & CSU Data1 & UNE & NUCES \\
        \hline\hline
        \multirow{5}{*}{YOLOv3 (Fine tuned)} 
        & 160 & 0.829 & 0.985 & 0.988 \\
        & 80  & 0.834 & 0.887 & 0.954 \\
        & 40  & \cellcolor{lightgreen}0.712 & \cellcolor{lightgreen}0.697 & \cellcolor{lightgreen}0.661 \\
        & 20  & \cellcolor{lightgreen}0.661 & \cellcolor{lightgreen}0.635 & \cellcolor{lightgreen}0.596 \\
        & 10  & \cellcolor{lightgreen}0.491 & \cellcolor{lightgreen}0.365 & \cellcolor{lightgreen}0.040 \\
        
        \midrule
        \multirow{5}{*}{YOLOv5 (Fine tuned)} 
        & 160 & 0.874 & 0.975 & 0.995 \\
        & 80  & 0.848 & 0.975 & 0.995 \\
        & 40  & \cellcolor{lightgreen}0.738 & \cellcolor{lightgreen}0.655 & \cellcolor{lightgreen}0.682 \\
        & 20  &\cellcolor{lightgreen}0.637 & \cellcolor{lightgreen}0.629 & \cellcolor{lightgreen}0.437 \\
        & 10  & \cellcolor{lightgreen}0.443 & \cellcolor{lightgreen}0.464 & \cellcolor{lightgreen}0.039 \\
        \midrule
        
        \multirow{5}{*}{YOLOv7 (Fine tuned)} 
        & 160 & 0.968 & 0.995 & 0.995 \\
        & 80  & 0.968 & 0.995 & 0.995 \\
        & 40  & 0.765 & \cellcolor{lightgreen}0.729 & \cellcolor{lightgreen}0.592 \\
        & 20  & \cellcolor{lightgreen}0.689 & \cellcolor{lightgreen}0.581 & \cellcolor{lightgreen}0.441 \\
        & 10  & \cellcolor{lightgreen}0.431 & \cellcolor{lightgreen}0.419 & \cellcolor{lightgreen}0.189 \\
        \midrule
        
        \multirow{5}{*}{YOLOv8 (Fine tuned)} 
        & 160 & 0.985 & 0.995 & 0.995 \\
        & 80  & 0.989 & 0.995 & 0.995 \\
        & 40  & \cellcolor{lightgreen}0.731 & \cellcolor{lightgreen}0.728 & \cellcolor{lightgreen}0.647 \\
        & 20  & \cellcolor{lightgreen}0.558 & \cellcolor{lightgreen}0.493 & \cellcolor{lightgreen}0.227 \\
        & 10  & \cellcolor{lightgreen}0.139 & \cellcolor{lightgreen}0.256 & \cellcolor{lightgreen}0.045 \\
        \midrule
        
        Grounding DINO (Zero-shot) & 0 & \textbf{0.753} & \textbf{0.789} & \textbf{0.758} \\
        \bottomrule
    \end{tabular}
    }
    \label{tab:model_comparison}
\end{table}
Results indicate that Grounding DINO achieves strong zero-shot performance, with mAP@0.5 scores of 0.753 (CSU Data1), 0.789 (UNE), and 0.758 (NUCES), without requiring any fine-tuning or training. In contrast, all YOLO variants require between 40 and 80 labeled training samples to achieve comparable performance levels, highlighting the effectiveness of Grounding DINO in low-data scenarios. 

Grounding DINO offers significant practical advantages. Fine-tuning YOLO-based models (YOLOv3, YOLOv5, YOLOv7, YOLOv8) demands a considerable amount of time, computational resources, and manual effort. The process involves dataset preparation, manual annotation, training on GPU-enabled systems, and extensive hyperparameter tuning. Furthermore, the process must be repeated for each target dataset, increasing both the time and cost. In contrast, Grounding DINO provides a cost-effective and scalable alternative, capable of producing competitive results without the need for any supervised training. Its ability to generalize across datasets without re-training makes it particularly suitable for applications with limited data availability or constrained computational budgets. 

\section{Discussion}
\label{discussion}
As presented in Table~\ref{tab:model_comparison}, the recent studies have utilized supervised deep learning models such as YOLOv3, YOLOv5, YOLOv7, and YOLOv8, achieving high accuracy, up to 99.5\% in some cases. However, a major limitation of supervised models is their dependency on the specific dataset they are trained on. These models are typically optimized for particular cattle breeds or controlled environments. If a new dataset is introduced or if the breed changes, the entire dataset must be annotated and retrained from scratch, which is resource-intensive and impractical for real-time or large-scale applications.

In contrast, the zero-shot detection method explored in this study using Grounding DINO does not require task-specific retraining. It leverages natural language prompts (e.g., ``cattle muzzle'' or ``snout of a cattle'') to detect the target object, making it more flexible and generalizable across different cattle breeds and unseen environments. Although the accuracy (76.8\%) is lower than that of supervised models, this trade-off is offset by the method’s adaptability, annotation-free deployment, and potential to scale across breeds and geographies. Therefore, zero-shot detection presents a promising direction for practical muzzle detection applications.

Future work can focus on further improving the accuracy of zero-shot muzzle detection through advanced vision-language models and refined prompt design. Using detailed, domain-specific prompts can lead to more accurate and reliable results. A promising direction is enabling natural language queries, such as ``How many cattle have a white muzzle?'' or ``What is the medication history of cattle with a white muzzle with red marks?'', to support intuitive interaction and enhance traceability. This would allow farmers and livestock managers to monitor and search for specific cattle using simple descriptions. Further evaluation across breeds and conditions, along with optimization for edge deployment, will support practical and scalable use in real-world farming environments. 

\section{Conclusion}
\label{conclusion}
In this study, we selected seven ZSD models capable of handling long and descriptive prompts to detect cattle muzzles. We explored various custom natural language prompts to describe muzzle characteristics and identified the most effective ones for detection performance. Through a comprehensive evaluation, we assessed the performance of all seven ZSD models and found that Grounding DINO outperformed the others in terms of accuracy and consistency. Unlike traditional supervised methods, Grounding DINO offers the advantage of detecting muzzles without requiring task-specific training data, making it more adaptable across breeds and environments. These findings demonstrate the potential of prompt-driven ZSD models for practical and scalable livestock monitoring. Nevertheless, certain limitations should be acknowledged. Large language models remain susceptible to hallucination, which may compromise the reliability of generated guidance. Furthermore, the approach is sensitive to prompt design, as variations in phrasing can influence both the consistency and quality of results.

\section*{Acknowledgment}
This project was supported by funding from Food Agility CRC Ltd, funded under the Commonwealth Government CRC Program. The CRC Program supports industry-led collaborations between industry, researchers, and the community. We thank the travel support from Gulbali Institute for Agriculture, Water and Environment. We also thank Dr Shawn McGrath, Mr. Jonathan Medway, and Prof. Dave Swain for their assistance with the project.
 \bibliographystyle{splncs04}
\bibliography{reference}

\end{document}